# Fast Leave-One-Out Approximation from Fragment–Target Prevalence Vectors (molFTP) : From Dummy Masking to Key-LOO for Leakage-Free Feature Construction

Guillaume Godin[1]

[1]Osmo Labs PBC New York, USA

Corresponding authors: guillaume@osmo.ai

**Abstract**

- We introduce molFTP (molecular fragment-target prevalence), a compact representation that delivers strong predictive performance. To prevent feature leakage across cross-validation folds, we implement a dummy-masking procedure that removes information about fragments present in the held-out molecules. We further show that key leave-one-out (key-loo) closely approximates true molecule-level leave-one-out (LOO), with deviation below 8% on our datasets. This enables near–full-data training while preserving unbiased cross-validation estimates of model performance. Overall, molFTP provides a fast, leakage-resistant fragment–target prevalence vectorization with practical safeguards (dummy masking or key-LOO) that approximate LOO at a fraction of its cost.

    **Scientific Contribution.** We provide a fast fragment-target prevalence vectorization (molFTP) coupled with dummy masking and key-LOO options, yielding a feature-leakage-free approximation to molecule-level LOO for robust cross-validation.

# Introduction

A fundamental question for users of predictive models is : how good is the training data (1) ? One way to approach this is by delineating the model's applicability domain. A second safeguard is to prevent data leakage (2), which motivates deduplication and proper validation protocols (3). In practice, it is standard to use cross-validation or time-series split such as SIMPD (4). Beyond sample-level leakage (molecules crossing folds), we must also consider feature leakage (when features inadvertently encode information about held-out molecules) (2). We return to this point in the related Work section.

In Classification, label ambiguity is common : even simple binary tasks (aka Cat vs Dog) exhibit non-zero label error rates (5). In pharmacology, practitioners may adopt a decision boundary (e.g., $1\times10^{-6}$ M) to separate "active" from "inactive" compounds (6). Because both experimental measurements and reported concentrations carry uncertainty, **x** (inputs) and **y** (labels) are both noisy (7). A pragmatic mitigation is to exclude molecules near the decision threshold by defining an indeterminate buffer zone (e.g., $>5\times10^{-6}$ M = inactive, $<5\times10^{-7}$ M = active, and $[5\times10^{-7}, 5\times10^{-6}]$ M = indeterminate). This reduces label noise and typically yields more reliable models.

Recent discussions on accuracy metrics for training LLMs raise an analogous issue (8) : boundary-adjacent points increase label noise and induce guessing, weakening models (9). Therefore, we need methods that surface problematic data. In our setting, using an unhashed fragment representation computed on (nearly) the full dataset preserves domain knowledge for generalization and helps gauge the magnitude of potential misclassifications. If a knowledge-based feature set cannot correctly predict a molecule's class, one should examine whether the cause is an applicability-domain mismatch or a labeling error.

When features are supervised, knowledge-aware, they are especially prone to data leakage (10).The classical fix is to recompute features within each fold of cross-validation. Complementarily, recent work shows that Approximate Leave-one-out (ALO) can provide fast, analytic approximations to LOO at training time (11,12). Here, we proposed a pipeline that extracts full supervised chemical knowledge without sacrificing data, while satisfying two constraints: leakage-free features and support for generalization. We introduce two strategies: (i) dummy masking, which removes contributions from fragments present in the held-out set, and (ii) key leave-one-out (key-LOO), which excludes singleton keys on a per-molecule basis (13). Neither requires recomputing global feature distributions for every split. Dummy masking requires knowledge of test indices and is practical for standard CV; key-LOO generalizes to the LOO extreme. With N molecules (N-1 train, 1 test), both methods are identical by construction. Moreover, key-LOO can be applied without pre-specifying test indices and approaches the asymptotic "true" LOO regime (N folds).

Finally, we define a molecule-level vectorization that captures per-radius fragment-target statistical enrichment using margin-aware aggregation functions. Applying dummy masking during cross-validation or key-LOO yields a feature-leakage-free representation, enabling unbiased estimation of model performance during ML cross-validation.

Our main contributions are the following:

1. **Fragment–Target Prevalence (FTP)**. We define per-radius fragment-target prevalence statistics.

2. **Higher-order FTP interactions**. We extend FTP to second- (pairs) and third-order (triplets) interactions, filtered by a similarity threshold.

3. **molFTP representation**. We map per-radius FTP keys, pairs, and triplets into a fixed-length molecular feature vector (parameterized by maximum radius) using margin-based aggregation functions of positives versus negatives fragment effects.

4. **Leakage control (Dummy Masking)**. We zero out test-only fragment contributions within each cross-validation fold to prevent feature leakage.

5. **Leakage control (Key-LOO)**. We introduce Key Leave-One-Out (key-loo) as a mathematically first approximation to true LOO that preserves knowledge while minimizing and controlling potential data leakage during prevalence estimation.

6. **Prediction with molFTP.** We train predictors directly on FTP-based molecular vectors, using leakage-control strategies.

7. **Generalization and ensembles.** We proposed to combine multiple molFTP as a model ensemble or concatenate molFTP with other features as a feature ensemble.

Fragment-target prevalence features enable strong predictive performance with controlled leakage and improved generalization.

# Data

Our Blood-Brain Barrier Penetration (BBBP) datasets were compiled from multiple sources (Table 1). Our goal is to train accurate models while explicitly accounting for variability in input data quality and noise. We also evaluate a second dataset that predicts a molecule's likelihood of being perceived as odorless. BBBP labels are derived from concentration measurements, whereas Odorless labels come from human-panel annotations and are therefore more susceptible to bias. We report Tanimoto similarity distributions for the two main datasets, BBB_free and Odorless, in figure 1, with expanded plots provided in the Supplementary Information (figure F.2).

| Dataset name | Observation | Size |
| --- | --- | --- |
| BBBP_free | First original paper source filtering unique molecules (14) | 1957 |
| B3DB | Highly curated dataset from multiple sources(15) | 7807 |
| BBBP4094 | Not fully curated dataset(16) | 4094 |
| BBBP4094sub | Overlap BBBP4094 with B3DB (labels BBBP4094) | 2937 |
| B3DBsub | Overlap B3DB with BBBP4094 (labels B3DB) | 2937 |
| BBBP_fix | Overlap with B3DB and BBBP_free using B3DB labels | 1721 |
| BBBP10%flip | BBBP_fix + flip 10% of labels | 1721 |
| Odorless | very diverse molecules in odorless part compar to the odor part see annexes part(17) | 6656 |

Table 1 : Database sources or manipulation

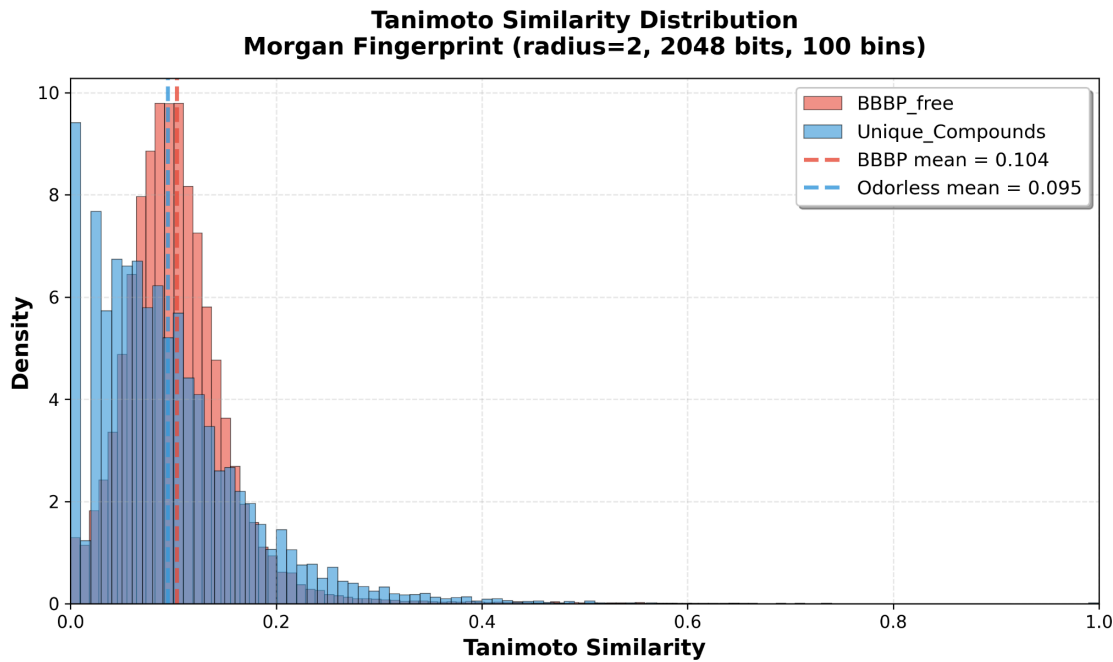

Figure 1: datasets similarity distribution of one example of BBBP dataset and the Odorless dataset.

## Relative works

Feature leakage is distinct from sample/training leakage. It arises when features encode prior knowledge about the target. Example using cLogP to predict logP/logD, which are closely related properties (18). Even absent causality, correlations (including those embedded in pretrained models) can induce leakage; this helps explain why combinations of Abraham descriptors often perform well on classical physicochemical tasks (19).

Another well known usage of supervised target-aware knowledge is also used in ligand binding pocket derived pharmacophores, which rely on 3D modeling and, frequently, crystal protein structures. Methods that could be accurate but computationally expensive and data-limited (20). Because datasets are small (<100 molecules), leave-one-out (LOO) is commonly used for validation, and large-scale virtual screening remains highly expensive and time consuming.

Recent neural approaches embed prior knowledge either explicitly such as molGPS a target-aware graph prior (21) or implicitly via target-agnostic pretraining, such as Transformer-CNN, which learns chemistry from augmented SMILES reconstruction without target labels (22). Similarly, MGTP extracts fingerprints with masked-graph transformer encoders, akin to Transformer-CNN's encoder stage (16).

Fragment target based analysis is not new; it has been used for applicability-domain assessment and molecular-field–based generative methods (23,24). The open question is how to provide efficient, feature leakage-free fragment-target features for modern machine learning.

# Method

We estimate fragment-target prevalence and derive, for each molecular fragment, its preference for the positive versus negative class. During cross-validation, we prevent feature leakage by dummy masking: setting to zero the contributions of fragments that are absent from the training fold. We compute all keys and statistics once on the full corpus, then re-normalize within each fold using only training-fold molecules (not the global set). We also propose Key Leave-One-Out (key-LOO), a limit-case strategy that approximates molecule-level LOO by removing the influence of singleton keys at the per-molecule level.

Using RDKit (25), we enumerate circular fragments for radii r = 0 to R for all molecules. For each fragment key, we build a 2×2 contingency table (presence/absence × class) to capture prevalence across targets. We compute 1D per-fragment prevalence statistics (Fisher's formulation of wk) over the full dataset. Storing raw fragment features, we apply statistical pooling to convert fragment-level statistics into a fixed-length molecular vector (see figure 2).
We extend the scheme to 2D (pairs) and 3D (triplets) (aka contrastive pairs (26) and activity cliff triplet(27)), using appropriate tests (McNemar (28) and Binomial (29) or Friedman (30) statistics respectively).

### 1D prevalence in molFTP

For each fragment $K$ with contingency counts $(a, b, c, d)$ and Haldane smoothing $\alpha = \frac{1}{2}$, the **1D prevalence** is the smoothed log-odds ratio

$$w_K = \log_2 \frac{(a+\alpha)(d+\alpha)}{(b+\alpha)(c+\alpha)}. \tag{1}$$

A Fisher-style significance score is derived directly from $w_K$:

$$\text{score} = \text{sgn}(w_K) \left[ -\log_{10}(\max(p, 10^{-300})) \right], \tag{2}$$

where $p = \text{erfc}(|w_K|/(\sqrt{2\,\widehat{\text{Var}}(w_K)}))$ with $\widehat{\text{Var}}(w_K) = \frac{1}{(\ln 2)^2}\left(\frac{1}{a+\alpha} + \frac{1}{b+\alpha} + \frac{1}{c+\alpha} + \frac{1}{d+\alpha}\right)$.
At inference, these weights are aggregated per molecule into the fixed-length vector

$$\boxed{\mathbf{V}^{(1D)} = [\,\text{margin}, \text{margin}_{\text{rel}}, \text{net}_0, \ldots, \text{net}_R\,] \in \mathbb{R}^{2+R+1}} \tag{5}$$

which is concatenated with analogous 2D/3D views to form the molFTP representation.

Figure 2: 1D prevalence vectorization from database fragment target scoring to vector creation

Within each CV fold, we zero out contributions from fragments not observed in the training fold and rescale counts on the training fold only using the Dummy masking method.

Alternatively, in key-loo, we remove the influence of keys observed in only one molecule (singletons), yielding a fast first-order approximation to true LOO that preserves knowledge while minimizing and controlling potential data leakage during prevalence estimation. Dummy masking requires fold indices; key-LOO can be computed without them and leverages nearly all available data.

The aggregating pooled FTP-based molecular vectors (per radius, including pair/triplet terms) serve as inputs to downstream ML models. We report results on BBBP datasets and on the Odorless dataset.

## MolFTP vector

We construct the molFTP vector with a max-based aggregation that maps per-molecule, per-radius FTP statistics to two scalars: a margin and a relative margin. The margin is defined as the maximum positive contribution minus the minimum negative contribution across fragment-level (atom-localized) scores, capturing the extreme separation; the relative margin is its normalized counterpart. Besides the max operator used for the margin, we also support alternative pooling operators (softmax, mean, median, log-sum-exp). Margins are computed separately for each FTP statistical type (figure 4), and the final molFTP vector is the concatenation of the three interaction orders : 1D (single keys), 2D (pairs), and 3D (triplets). Vectors can be generated using either binary presence or count-based occurrence, as in ECFP (31). To clarify this distinction, consider trinitrotoluene (TNT): it is not the mere presence of a nitro fragment that yields explosive behavior (mononitrotoluenes are widely used in pigments, antioxidants, and agrochemical), but the count of three nitro groups on toluene. In practice, for molFTP we typically observe little difference between counting and binary presence because aggregation operates on statistical scores rather than raw counts (figure 3), akin to pooling in a graph neural network.

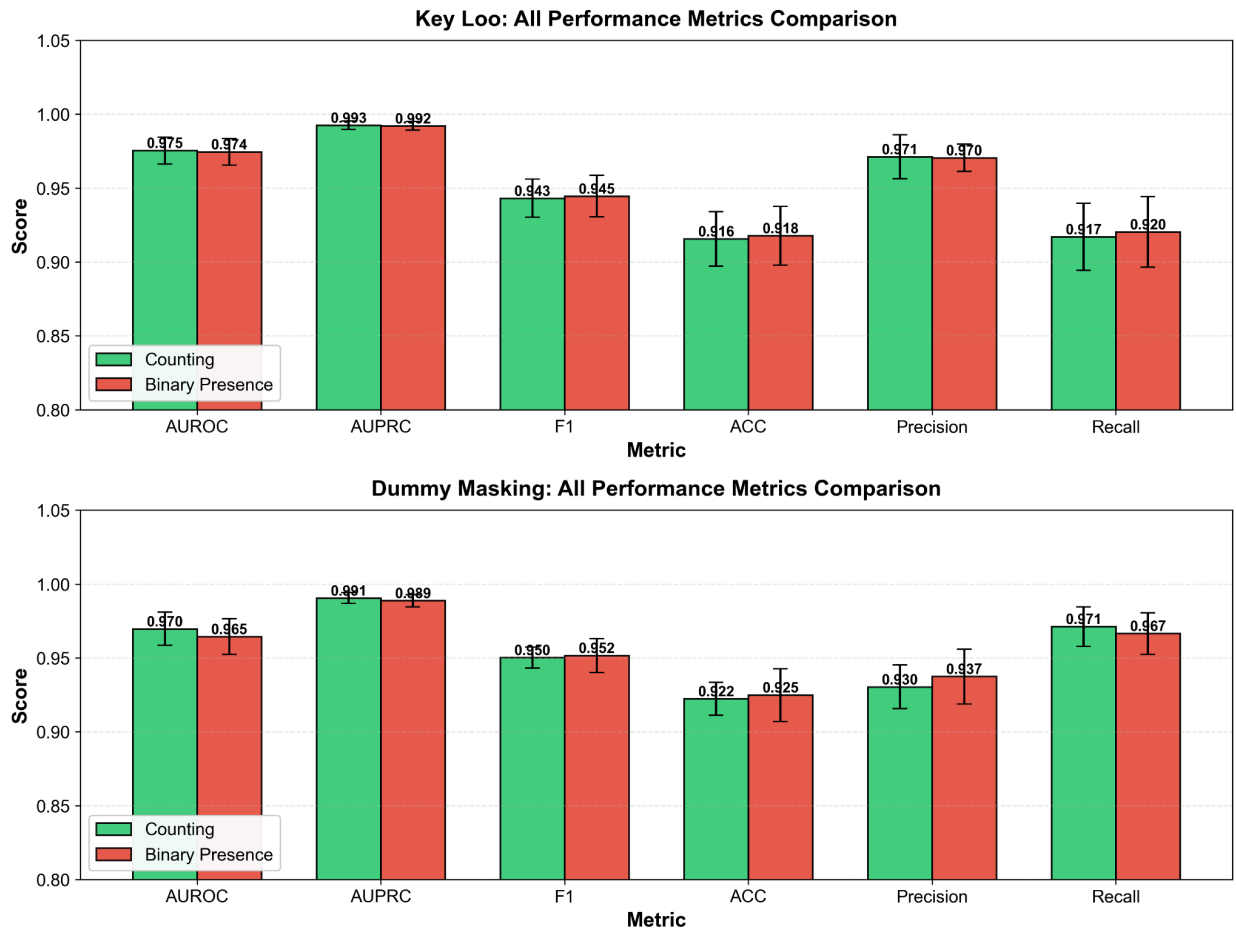

Figure 3: No visible difference on Counting versus Presence statistical effect on molFTP on both dummy masking and key-loo. We observed that Precision is better for Key-LOO and Recall for Dummy masking.

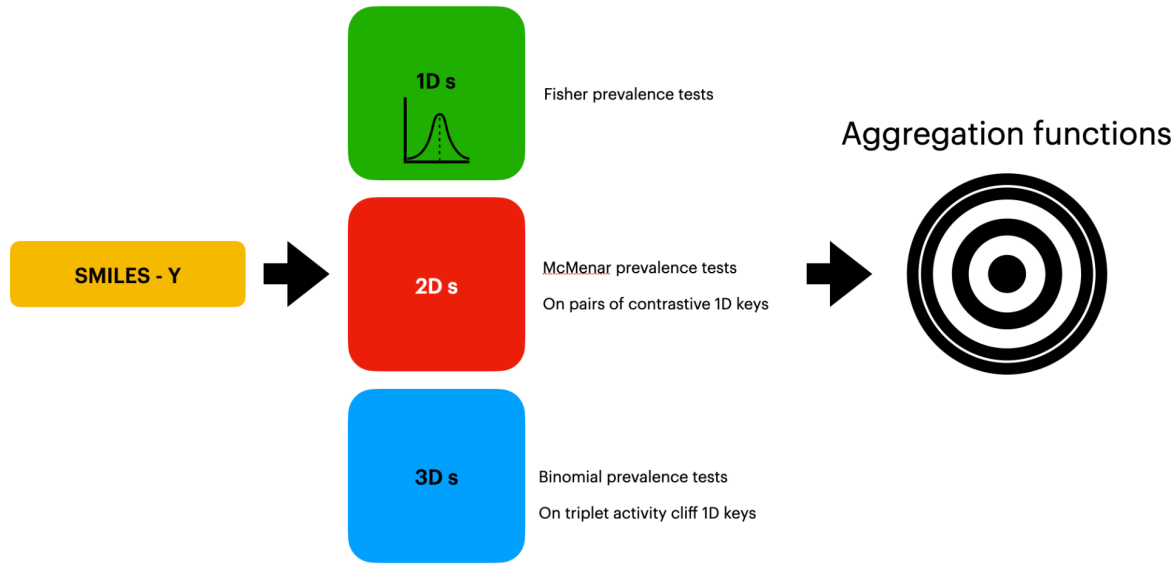

Figure 4: Statistical target keys prevalences single, pair and triplet keys conversion into a molecule knowledge vector

# Results

We evaluated molFTP on the BBBP collections previously introduced in (32) originally used to illustrate bond-centered molecular fingerprint derivatives. Unless stated otherwise, we set the maximum radius to R = 6 and the similarity threshold to tau = 0.5 when constructing higher-order meta-keys (pairs and triplets). Models were trained with 10-fold cross-validation using logistic regression (33) or random forest (34) implemented in scikit-learn (35). Figure 5 reports performance for both leakage-control strategies, dummy masking and key-LOO, under the default configuration ( R=6 ), which yields a 27-dimensional molFTP vector per molecule assembled from the 1D, 2D, and 3D features.

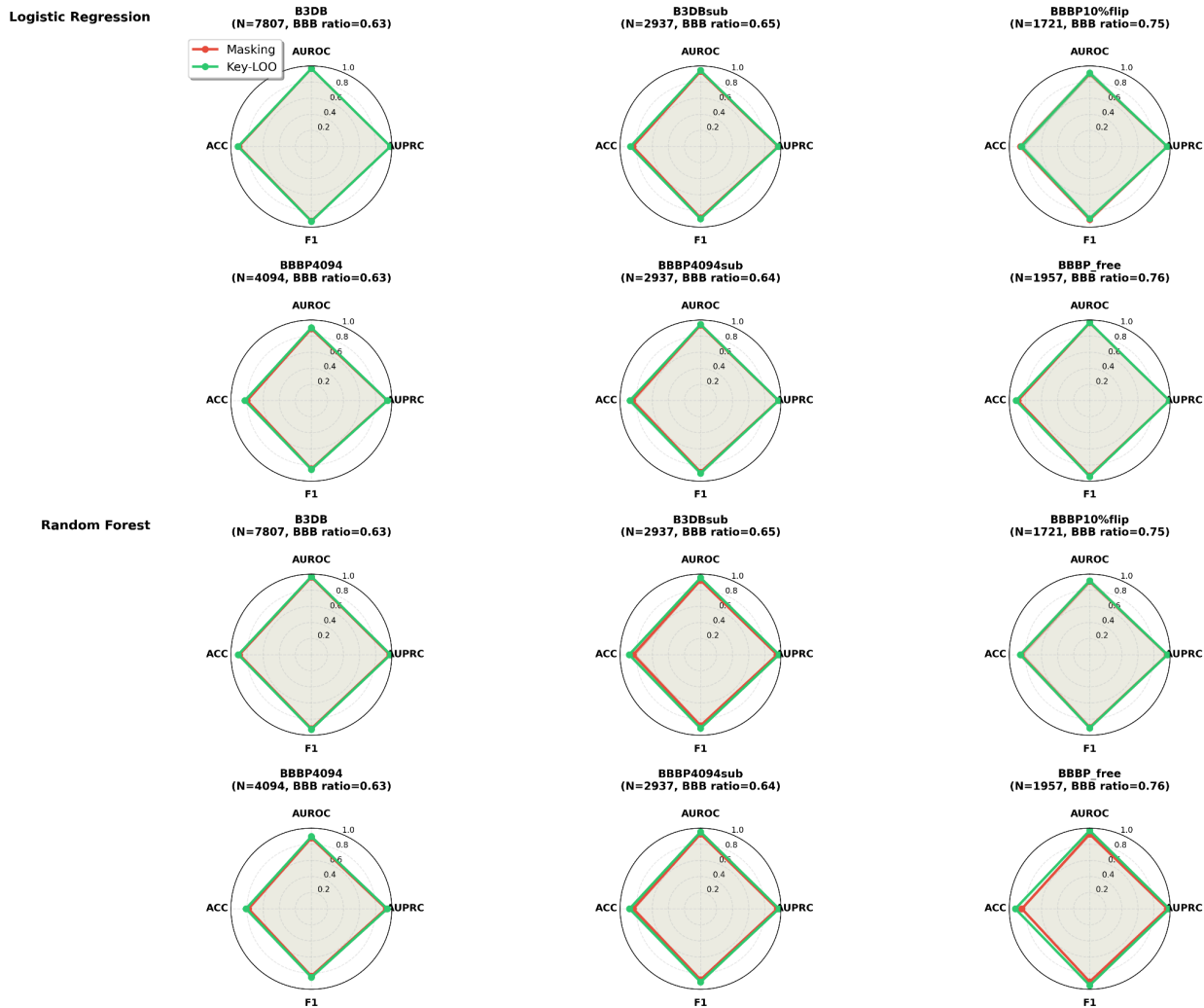

Figure 5: Summary of the model 10 CV performance using masking or key-loo principles, with radius 6 and similarity threshold 0.5 logistic regression or random forest

Table 2 focuses on the BBBP4094 dataset from figure 5 to compare against our prior baselines, namely "ECFP || BCFP (Sort&Slice + out-of-vocabulary bucket)" (32) and MGTP (16). While the E/BCFP_ss_oov baseline is generally on par with or slightly better than MGTP across the reported metrics, molFTP delivers consistent gains: with random forests, we observe absolute improvements of ~1–3% across both metrics. With XGBoost (36), the two molFTP methods become even closer than with Random Forest, reinforcing that key-LOO provides a first-order approximation to LOO and it is data-leak free.

| Molecular Features | Model | AUPRC | AUROC |
| --- | --- | --- | --- |
| MGTP | XGBoost | 0.9115+/-0.0155 | 0.8638+/-0.0242 |
| MGTP | Random Forest | 0.9016±0.0183 | 0.8495±0.0266 |
| E/BCFP_OOV_ss concat | XGBoost | 0.9163±0.0129 (r0) | 0.8636±0.0186 (r0) |
| E/BCFP_OOV_ss concat / hybrid | Random Forest | 0.9050±0.0125 (r1) | 0.8427±0.1967 (r2) |
| MolFTP-key-loo | Random Forest | *0.9441±0.0065* | *0.8995±0.0127* |
| MolFTP-dummy-mask | Random Forest | *0.9350±0.0092* | *0.8830±0.0160* |
| MolFTP-key-loo | XGBoost | **0.9490±0.0058** | **0.9053±0.0109** |
| MolFTP-dummy-mask | XGBoost | *0.9444±0.0077* | *0.8932±0.0146* |

Table 2: E/BCFP_OOV_ss, MGTP and MolFTP comparison using a 10 CV for Random Forest or XGBoost (in bold best model, in italic closest model)

## Dummy Masking method

Our first leakage-control strategy alters fragment–target prevalence only marginally (mean absolute shift < 0.2), showing we can suppress leakage while preserving the statistics needed for accurate property prediction. For a test molecule, we zero the contribution of any FTP key not observed in the training fold (unseen keys). For keys observed globally but only partially within the fold, we need to downweight the global statistic. We adjust the FTP scores by applying a factor correction as the number of train molecules having this FTP divided by the total number molecules having this FTP in the full dataset. This factor-correction removes influence from test-set occurrences of "seen" keys without per-fold recomputation. Figure 6 plots the log-scaled difference between the global FTP distribution and the masked + corrected training-fold distribution.

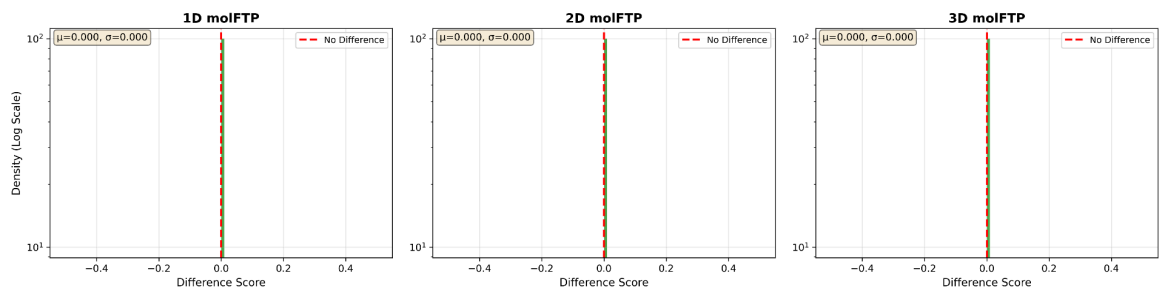
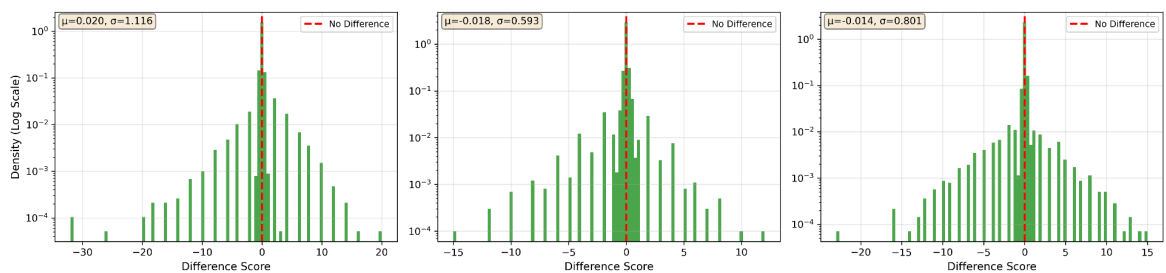
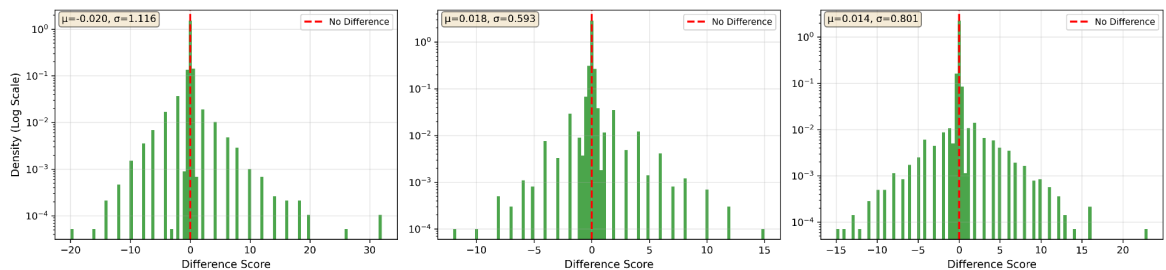

Figure 6: Dummy masking ablation versus key-loo ablation on proportions in 1D, 2D and 3D vectors

# Key leave one out first order approximation

In the supplementary part (demonstration D.1 and figure F.1), we empirically validate that key-LOO is a first-order approximation to molecule-level LOO. Using Eq. A.5 to define the permissible first-order LOO bound, we measure, for each dataset, the fraction of feature scores that fall within this bound. Across BBBP_free and Odorless, 92–94% of features lie within the bound; <8% exceed it, supporting key-LOO as a practical first-order approximation to true LOO.

# Single, Pair and Triplet keys statistical methods

We incorporate pair (2D) contrastive keys and triplet (3D) anchor keys to inject activity-cliff signal, using McNemar's test (pairs) and binomial/Friedman tests (triplets), with a similarity threshold that restricts analysis to closely related structures. We also implemented additional contingency/enrichment tests (Table 3), enabling alternative FTP score formulations and corresponding molFTP vectors. Empirically, we obtain the best performance by concatenating the single (1D), pair (2D), and triplet (3D) vectors (figure 7). The pair+triplet combination performs as well as or better than the single 1D vector alone, although this advantage diminishes at very strict similarity cutoffs. We further observe reduced variance across 10-fold CV when using single+pair+triplet (and similarly with single+triplet), consistent with triplets effectively capturing activity-cliff information. Notably, 1D alone is insufficient; 2D and/or 3D features generally improve model inputs.

| Statistical Test Type | Names | Usage | Best | Type |
|---|---|---|---|---|
| Proposed baseline | **Fisher one ot two tails** (1D) (37), **McNemar** (2D) (28), **Binomial** (3D) (29) | Contingency table | 1 | all |
| Asymptotic | Chi^2 (38), G-test, Z-test | large samples | 2 | 1D |
| Bayesian | beta-binomial | enrichment with priors | 2 | 1D |
| Information theory | PMI, NPMI, MI | association strength | 2 | 1D |
| Confidence interval | Wilson, Agresti-Couli | proportions | 3 | 1D,3D |
| Regularization | Shrunk estimates, Ridge | overfitting prevention | 2 | 3D |
| Contingency + | Barnard, Boschloo, CMH | better power | 2 | 1D |
| Other Pairs | Conditional LR | paired observations | 2 | 2D |
| Concordance | Cochran's Q, **Friedman** (30) | multi-class agreement | 1 | 3D |

Table 3 : List of statistics evaluated in **bold** the best one (**Friedman** is a little better than binomial for 3D)

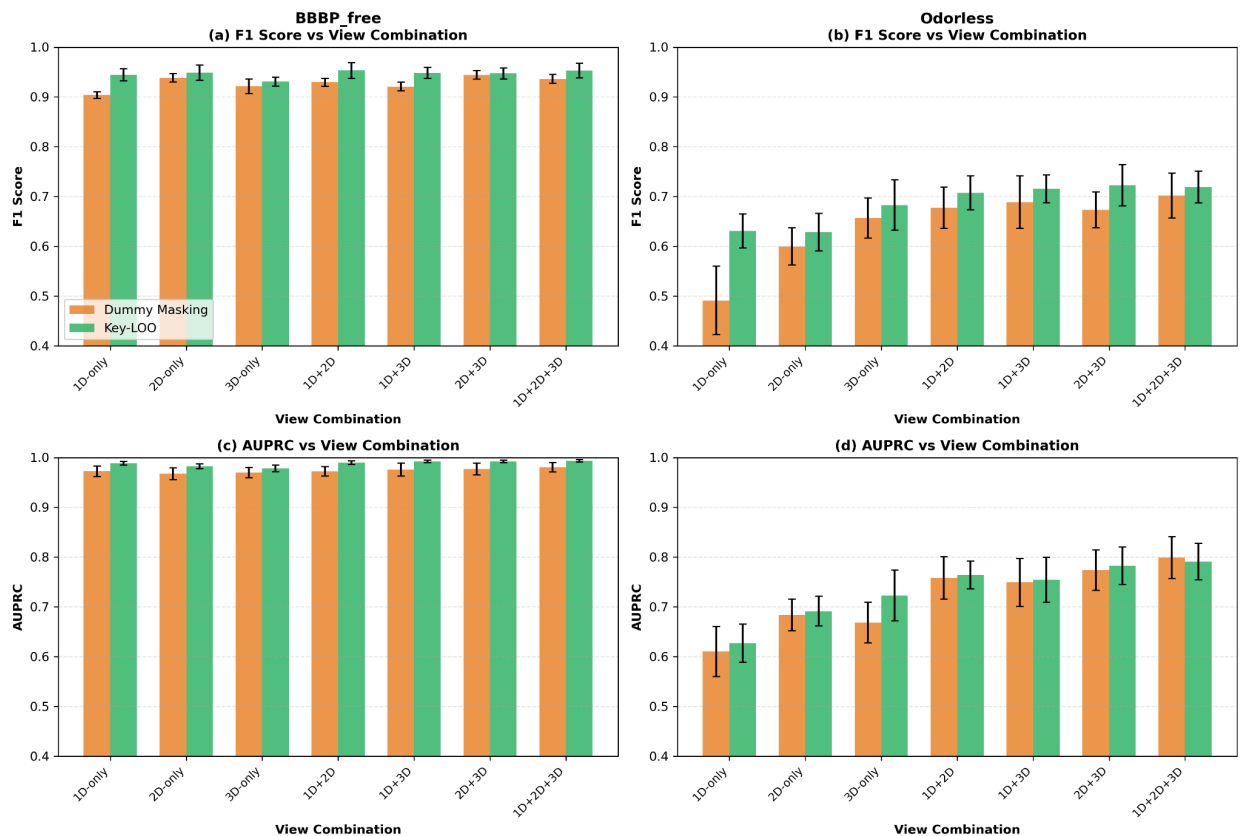

Figure 7: Study molFTP vector combination (baseline proposal) using 10 CV, radius 6, similarity threshold 0.5, using Key-LOO on a random forest model

For radius R = 6, each order yields a 9-dimensional vector (total 27-D when concatenated), which, to our knowledge, is among the most compact molecular encodings reported, aside from Abraham's six parameters (19) Osmordred features (39) that are typically tailored to physicochemical tasks such as solubility and logP. Sensitivity to the specific statistical test is modest (figure 8); accordingly, we use Fisher's one-tailed test for singles and McNemar plus binomial/Friedman for pairs/triplets as our baselines. We observe only minor performance differences across statistical choices (figure 8). Accordingly, we use Fisher's exact (one-tailed) for single-key features, McNemar's test for pairs, and the binomial test (with Friedman as an alternative) for triplets as our baselines.

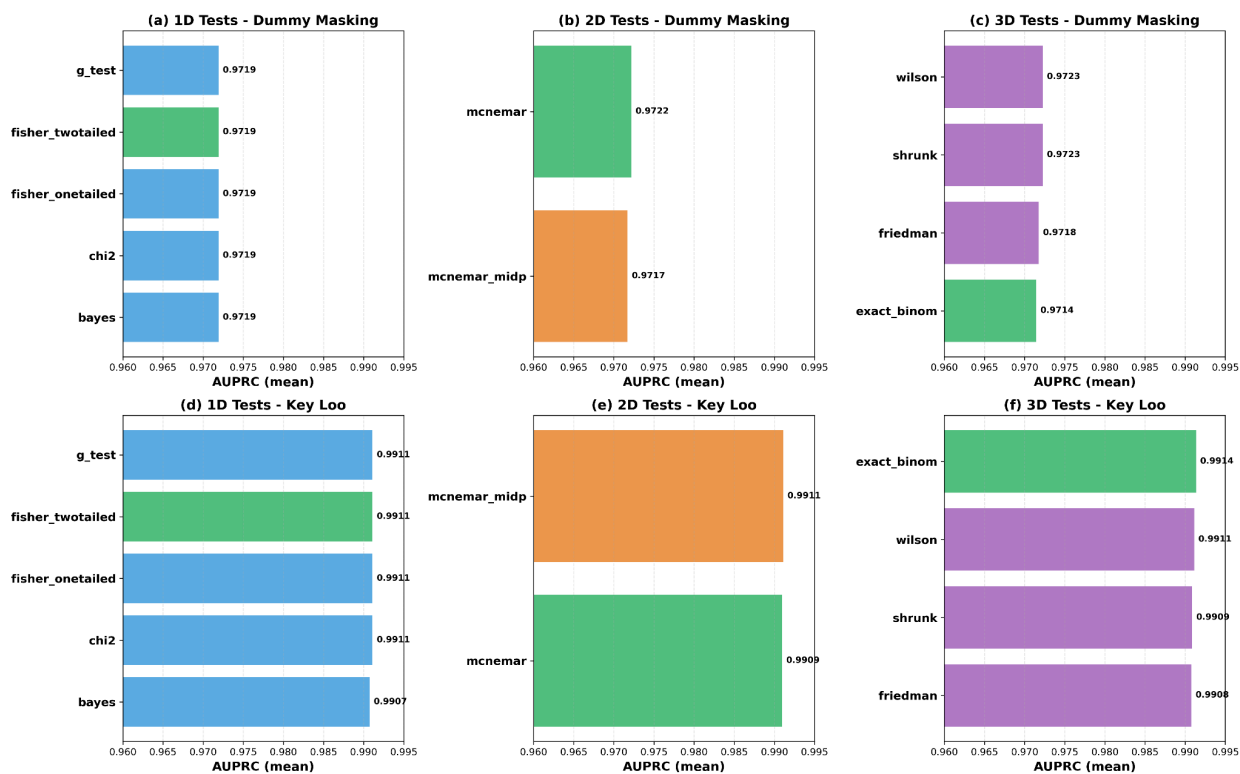

Figure 8: statistical effect on models compared few statistics analysis on BBBP_free dataset (Fisher, McNemar, Binomial and few potential 2D or 3D statistics).

# Aggregation margin functions

We pool FTP scores into the molFTP vector using several margin functions, trading off raw performance against generalization. While max pooling often delivers the strongest in-fold scores, it can over-emphasize extremes and scale poorly for harder generalization. Alternatives such as median, softmax, and log-sum-exp, tend to generalize better with only slight degradation in peak metrics. To highlight our two leakage-free strategies, we compare them with train-only molFTP prevalence vectors as classical cross validation of features. Aggregation choice has very small effects on BBBP and small-to-moderate effects on Odorless (see the Train-only columns). As expected, median and softmax yield very slightly higher AUROC/AUPRC than max under generalization stress. There is no universally best margin: Max is not always superior. Note that train-only 10-fold CV uses ~10% less data per fold than dummy masking or key-LOO, explaining its lower performance; it remains informative about generalization. By construction, LOO is the leakage-free gold standard, which key-LOO closely approximates. Results on Odorless (figure 9) mirror BBBP: the task is harder, max + train-only performs worse, and key-LOO and dummy masking are very similar, consistent with a more complex target.

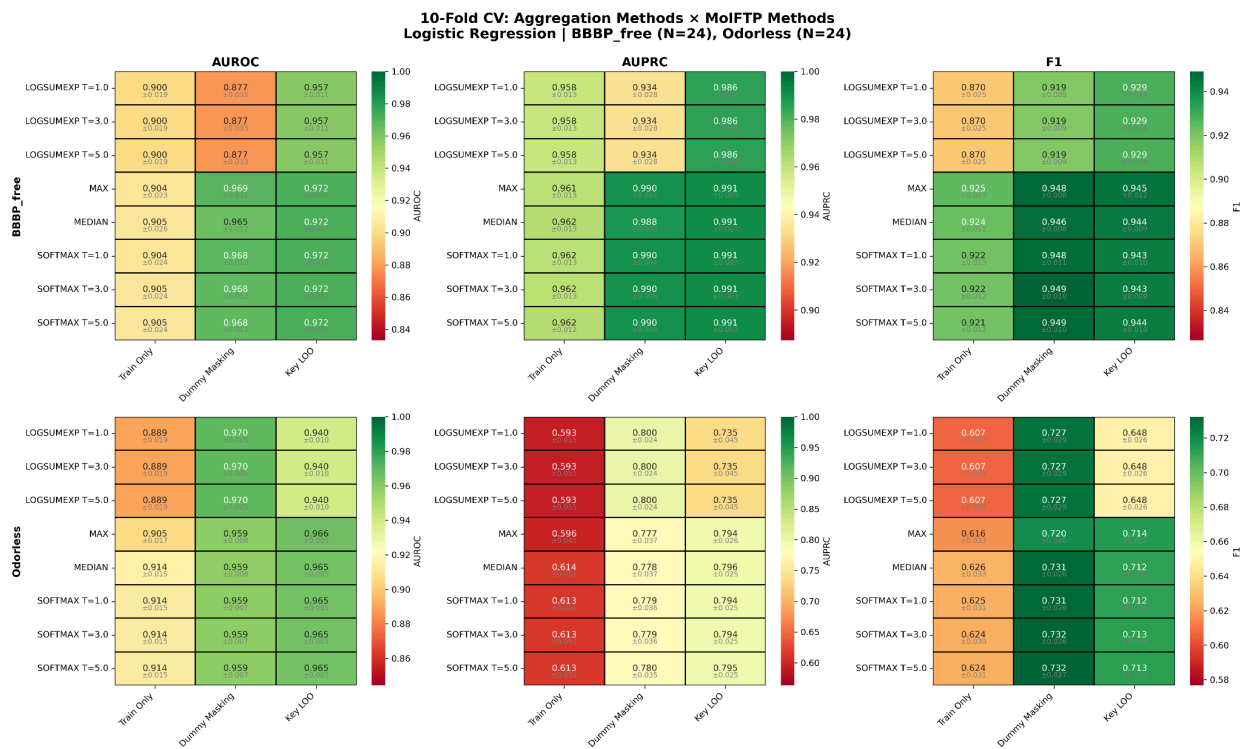

Figure 9: BBBP or Odorless free dataset using different aggregations functions (Max, Median, Logsumexp and Softmax)

A prior report combining ECFP || BCFP with random forests achieved F1 = 0.937 / AUC = 0.943 in 5×5 CV, closely matching our dummy masking / key-LOO performance in figure 9, despite using only 27 features with our molFTP compared to 4096 in the E/BCFP concatenated method (32).

# Radius factor effect

Tuning the radius adapts the representation to the target. Gains taper at larger radii (figure 10), yet random forests continue to benefit from the longer vectors. As expected, random forests outperform logistic regression across radii. The differences of Logistic regression and random forest is higher for Odorless metrics than BBBP_free dataset.

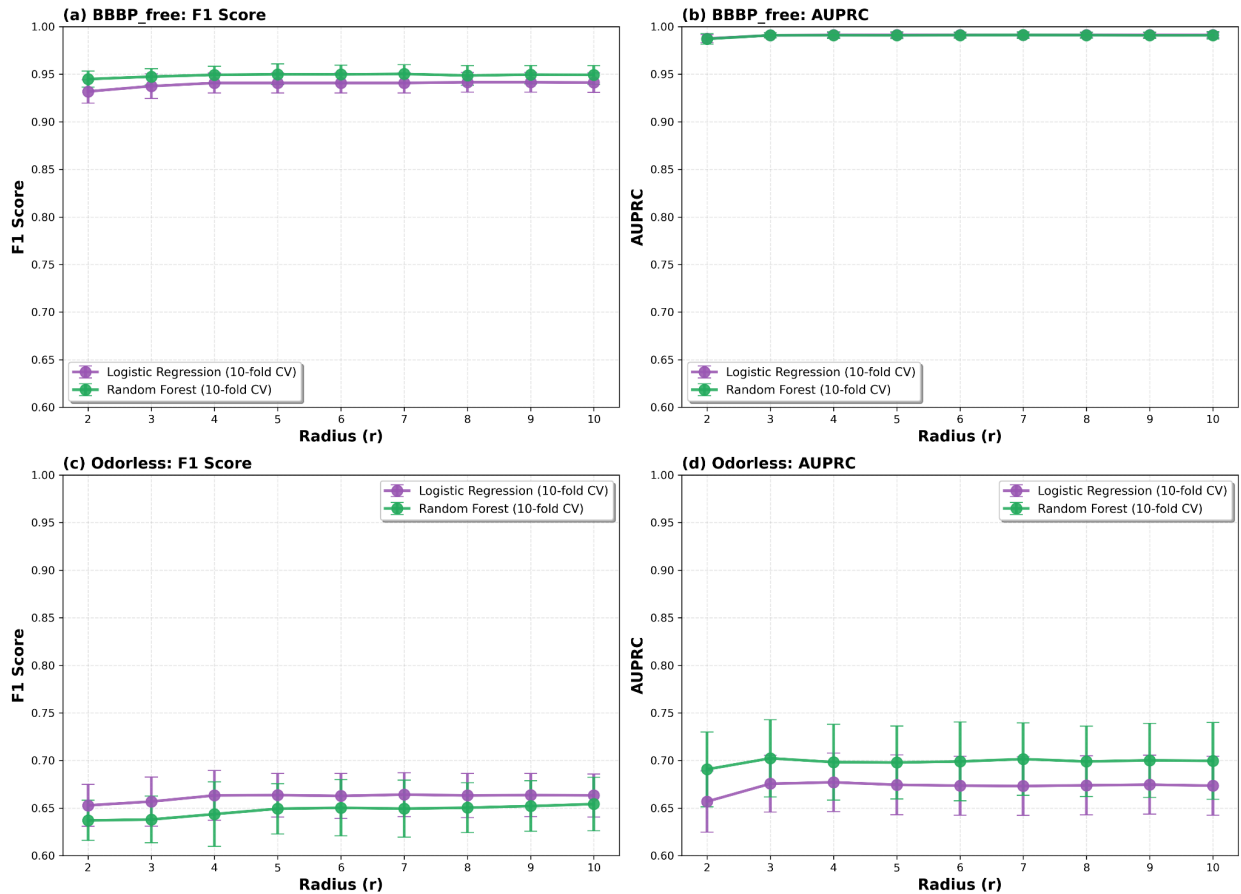

Figure 10: Radius molFTP study on Odorless and BBBP targets at a similarity threshold of 0.5 (10 CV for random forest or Logistic regression).

# Similarity Threshold factor effect

The optimal similarity threshold is dataset-dependent, as it governs which pair (2D) and triplet (3D) keys are included, especially important for challenging distributions like Odorless, whose Tanimoto similarities skew lower (figures 1, F.2). For Odorless, a threshold of 0.1 performs best (figure 11), and AUPRC approaches ~80%, underscoring the strong contribution of 2D/3D vectors discussed above.

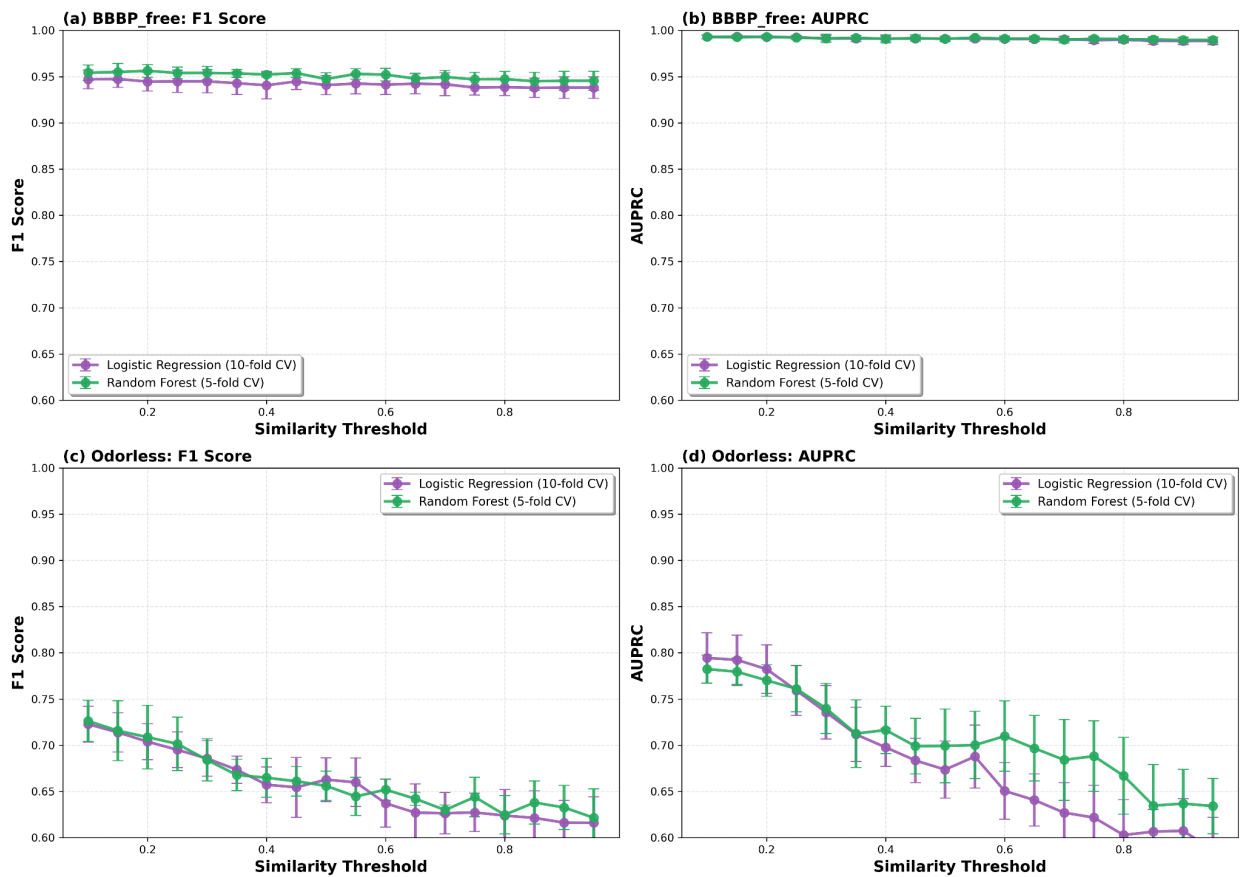

Figure 11: Similarity molFTP study on Odorless and BBBP targets at a radius of 6 (10 CV)

# Ensemble (model or feature)

molFTP naturally supports feature and model ensembling by combining multiple statistics and aggregation functions to produce diverse vectors. As a simple feature-ensemble test on Odorless, concatenating "sqrt(Molecular Weight), sqrt(Number of Atoms), sqrt(Number of Carbon Atoms)" to molFTP yields a clear lift, particularly with random forests (figure F.4). We expand on this in the Discussion. Similar ensemble gains have been observed previously on BBBP in related work (32).

# Discussion

Training on the full dataset yields an almost perfect in-sample model, aside from a few outliers we attribute to ambiguous labels, indicating strong self-knowledge extraction but also clear overfitting in the absence of an independent test set. To address this, we introduced two leakage-control strategies. Dummy masking zeros fragment–target preferences for fragments unseen in the training fold, disconnecting them from held-out targets. Key-LOO approximates molecule-level leave-one-out by removing singleton keys; it often (but not always) slightly outperforms dummy masking and can also introduce small biases. Using both methods provides a practical performance envelope.

Label quality strongly affects outcomes. In BBBP_free, 42 molecules carry labels that disagree with B3DB among the 1721 shared molecules. Our AUPRC is near-perfect on BBBP_free but notably lower on BBBP4094, consistent with higher noise being amplified by molFTP (figure 5). To probe sensitivity, we created BBBP10%flip by randomly flipping 10% of labels in BBBP_fix; performance degrades as expected (figure 12). In a second analysis, we found 2937 molecules common to BBBP4094 and B3DB with only 22 label mismatches; models trained on the BBBP4094 subset (BBBP4094sub) underperform those on the B3DB subset (B3DBsub), suggesting BBBP4094sub is noisier (figure 5).

Model capacity and vector expressivity also matter. Prior work showed our earlier best BBBP models trail those reported here, and performance trends upward with larger radius, plausibly because 2D/3D terms gain expressivity and vector length increases.

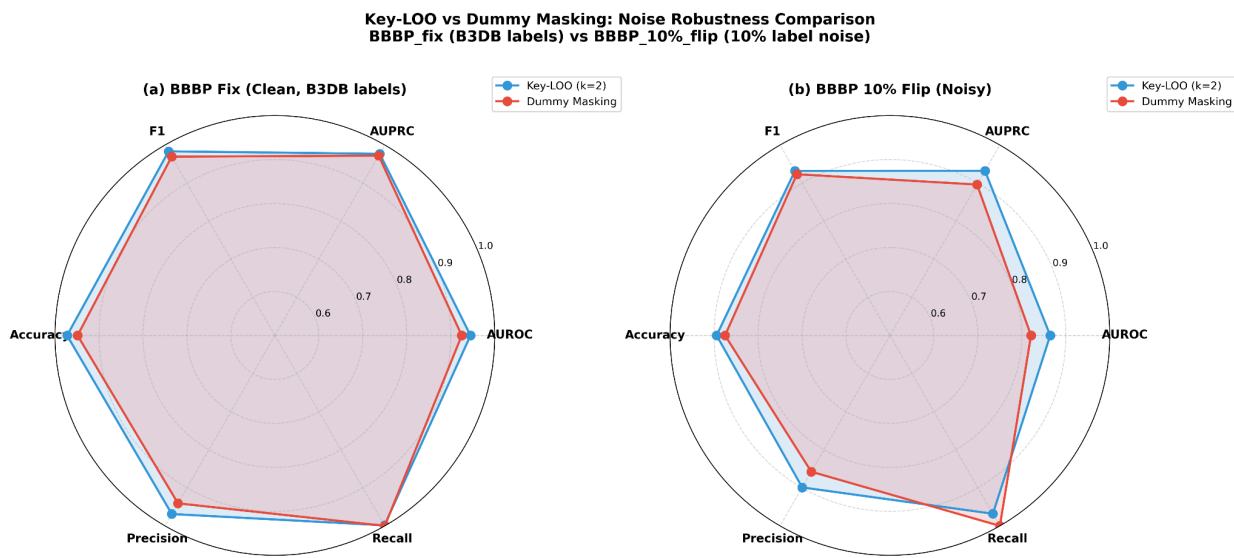

Figure 12: Adding noise label in BBBP dataset effect, Recall is almost preserve, while all the other metrics decrease as expected as we propagate noise in molFTP

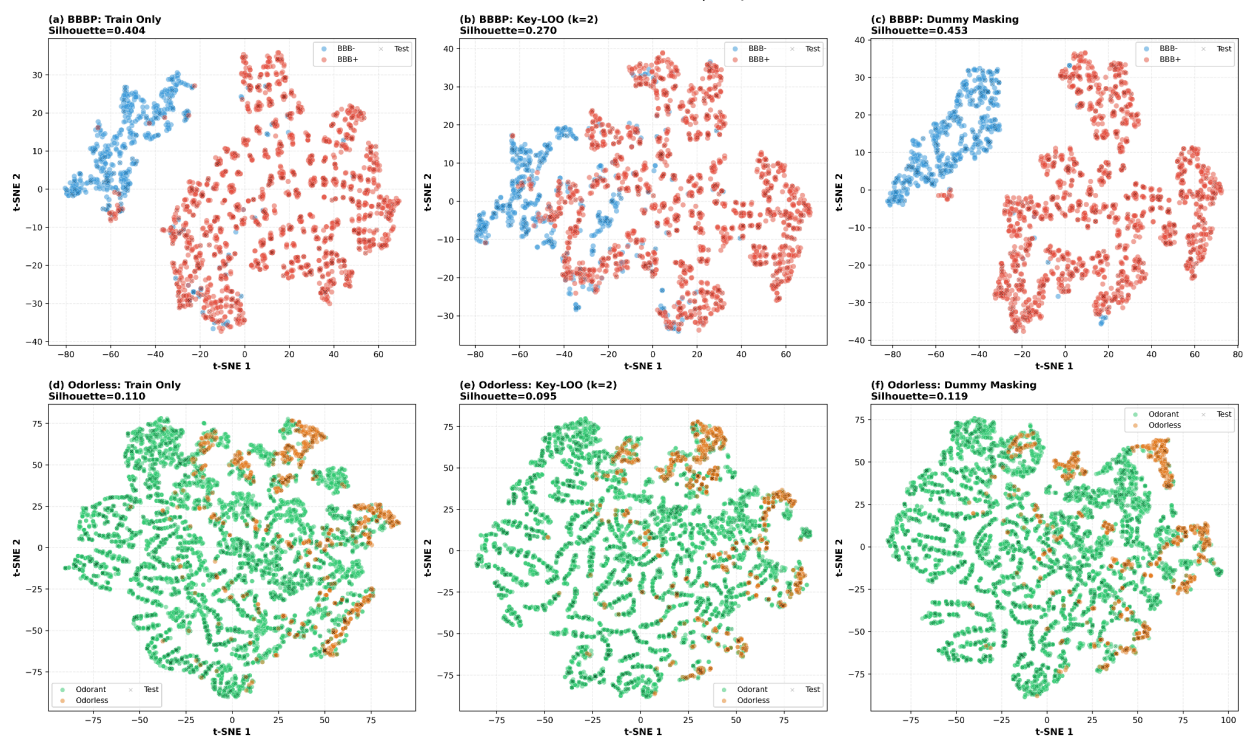

Figure 13: T-SNE BBBP_free and Odorless projections of the key-loo and masking compare to a train/test true split

Model capacity and vector expressivity also matter. Prior work showed our earlier best BBBP models trail those reported here, and performance trends upward with **larger radius**, plausibly because 2D/3D terms gain expressivity and vector length increases.

TNSE (40) or UMAP (41) projections of molFTP vectors (train-only vs. key-LOO vs. dummy masking) show BBBP classes are almost perfectly separated, especially with dummy masking, whereas Odorless exhibits less clearer separation (figures 13 and F.6). Dummy masking yields a higher silhouette score on BBBP, consistent with residual label issues; differences are small on Odorless, reflecting the task's difficulty. TSNE and UMAP thus offer didactic diagnostics for data quality and ambiguity captured by molFTP, helping flag challenging points influenced by weak or mispropagated target knowledge.

One might argue that supervised molFTP (dummy masking or key-LOO) still embeds target information. Empirically, the two-stage scoring and pooling compresses >100,000 key scores into a 27-dimensional vector at radius 6, which behaves more like strong compression / obfuscation than direct exposure. In Supplementary figure F.4, a random forest reaches AUPRC 0.87 with three simple molecular covariates appended and 0.84 without; logistic regression reaches 0.78 and 0.75, respectively, well below any "trivially supervised" ceiling, supporting that molFTP does not leak target labels in practice. Notably, dummy masking is close to key-LOO and surpasses ECFP (count/binary) baselines; Table 2 corroborates these findings too.

Finally, computational performance is favorable: on an M4 Pro MacBook (24 GB RAM), we generate ~2,000 molFTP vectors/s with RDKit and multithreading (Supplementary figure F.5), enabling thorough studies of label quality and first-order LOO approximation. The pipeline is summarized in Supplementary Schema F.3.

# Conclusion

We demonstrated that **m**olFTP features, coupled with dummy masking or key-LOO, deliver leakage-controlled performance, with key-LOO serving as a practical first-order approximation to molecule-level LOO and producing models closely aligned with dummy masking. molFTP achieves high compression efficiency: a vector of length 3 x (3+R) (e.g., 27 at R=6), among the most compact molecular projections we are aware of.

We quantified the influence of six design factors : radius, similarity threshold (for 2D/3D), statistical scoring, aggregation function, counting vs. presence, and feature selection. We showed how different choices yield complementary solutions suitable for ensembles. Augmenting molFTP with a few simple molecular descriptors can provide small but consistent gains, consistent with molFTP acting as a leakage-resistant compression of supervised knowledge rather than a direct channel for leakage.

Because the framework is grounded in explicit statistics and pooling, it is naturally differentiable and could be ported into neural architectures (e.g., as a learnable module or LM promptable component), an interesting direction for future work. As for any Machine learning model, a key limitation is label uncertainty : tasks relying on human panels (e.g., Odorless) are harder, as illustrated by the BBBP10%flip stress test. In turn, molFTP is useful for rapidly auditing data quality and triaging suspect labels. A multiclass molFTP extension, by vectorizing class-wise prevalence, has been implemented and shows robust results, but is beyond this paper's scope.

# Funding

The author was funded by Osmo Labs PBC for fragment data analysis and data quality analysis.

# Competing Interests and Consent for publication

The author declares that he has no competing interests. Author has read and agreed to the published version of the manuscript.

# Supplementary part

## Appendix A — Key-LOO as a First-Order Simulator of True LOO

**A.1 Notation and 1D prevalence.** Let $\mathcal{D} = \{(x_i, y_i)\}_{i=1}^{N}$ with $y_i \in \{0,1\}$. For key $K$, $X_{iK} \in \{0,1\}$. Define $a = \sum_i \mathbf{1}\{y_i = 1, X_{iK} = 1\}$, $b = \sum_i \mathbf{1}\{y_i = 0, X_{iK} = 1\}$, $c = \sum_i \mathbf{1}\{y_i = 1, X_{iK} = 0\}$, $d = \sum_i \mathbf{1}\{y_i = 0, X_{iK} = 0\}$, $N = a+b+c+d$. With Haldane $\alpha > 0$,

$$w(a,b,c,d) = \log_2 \frac{(a+\alpha)(d+\alpha)}{(b+\alpha)(c+\alpha)}. \tag{A.1}$$

**A.2 True LOO at the fragment level.**

$$\bar{w}_{\text{LOO}} = \frac{a}{N}w(a-1,b,c,d) + \frac{b}{N}w(a,b-1,c,d) + \frac{c}{N}w(a,b,c-1,d) + \frac{d}{N}w(a,b,c,d-1). \tag{A.2}$$

**A.3 Key-LOO simulator.**

$$w_{\text{KLOO}}(K) = \begin{cases} 0, & a+b < k, \\ s\, w(a,b,c,d), & a+b \geq k . \end{cases} \tag{A.3}$$

**A.4 LOO influence bound.**

$$\Delta_a = w(a-1,b,c,d) - w(a,b,c,d) = \log_2 \frac{a-1+\alpha}{a+\alpha} - \log_2 \frac{c-1+\alpha}{c+\alpha}. \tag{A.4}$$

**A.5 Approximation guarantee.**

$$\left| w_{\text{KLOO}} - \bar{w}_{\text{LOO}} \right| \leq \frac{C_\alpha}{k} + \left| s - \frac{N-1}{N} \right| \cdot \left| w(a,b,c,d) \right|. \tag{A.5}$$

Demonstration D.1 : Key-LOO First order simulator of True LOO

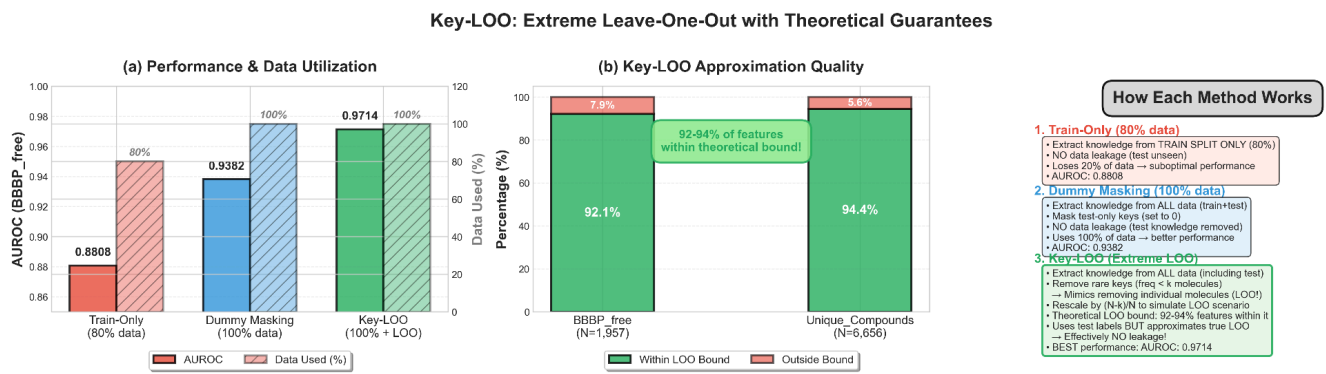

Figure F.1 : Illustration of the mask and key-loo methods and validity of the approximation equation A.5

## STEP 1 : from 1D prevalence score to molFTP vector

For each fragment (key) $K$, we form the *observational* $2 \times 2$ table

|      | with $K$ | without $K$ |
|------|----------|-------------|
| PASS | $a$      | $c$         |
| FAIL | $b$      | $d$         |

and apply Haldane smoothing $\alpha > 0$ ($\alpha = \frac{1}{2}$). The **1D prevalence** weight is the smoothed log-odds ratio

$$w_K = \log_2 \frac{(a+\alpha)(d+\alpha)}{(b+\alpha)(c+\alpha)}. \tag{1}$$

**Fisher-style significance.** Using $w_K$ directly, the variance estimate is

$$\widehat{\mathrm{Var}}(w_K) = \frac{1}{(\ln 2)^2}\left(\frac{1}{a+\alpha} + \frac{1}{b+\alpha} + \frac{1}{c+\alpha} + \frac{1}{d+\alpha}\right).$$

The standardized test statistic, $p$-value, and signed score are

$$z = \frac{|w_K|}{\sqrt{\widehat{\mathrm{Var}}(w_K)}}, \tag{1}$$

$$p = \mathrm{erfc}\left(\frac{z}{\sqrt{2}}\right), \tag{2}$$

$$\mathrm{score} = \mathrm{sgn}(w_K)\left[-\log_{10}(\max(p, 10^{-300}))\right]. \tag{2}$$

**molFTP embedding.** At inference, prevalence maps $\mathcal{E}_{\mathrm{PASS}}, \mathcal{E}_{\mathrm{FAIL}}$ assign $w_K$ to each hit $(K, d, a)$ (key, depth, atom). Per-atom scores are aggregated with gate $g$ (default 0) to yield

$$\mathrm{margin} = \#\{a : s_a \geq g\} - \#\{a : s_a \leq -g\}, \quad \mathrm{margin}_{\mathrm{rel}} = \frac{\mathrm{margin}}{|\mathcal{A}|}, \tag{3}$$

$$\mathrm{net}_d = \frac{\#\{a : s_{a,d} \geq g\} - \#\{a : s_{a,d} \leq -g\}}{|\mathcal{A}|}, \quad d = 0, \ldots, R. \tag{4}$$

Thus each molecule is represented by the fixed-length vector

$$\boxed{\mathbf{V}^{(1D)} = [\,\mathrm{margin}, \mathrm{margin}_{\mathrm{rel}}, \mathrm{net}_0, \ldots, \mathrm{net}_R\,] \in \mathbb{R}^{2+R+1}} \tag{5}$$

schema S.1 : prevalence vectorization

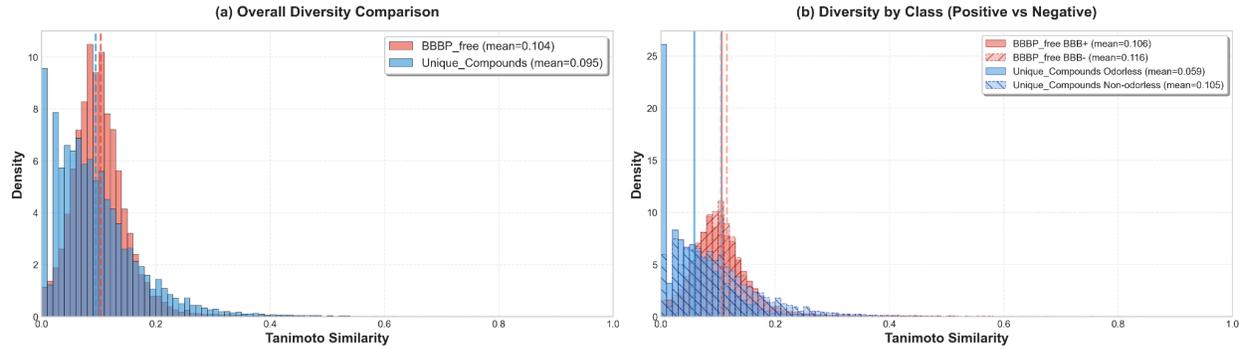

Figure F.2 : positive versus negative distributions of both dataset. Odorless and non-Odorless density distributions are very different. while this is not the case for BBBP.

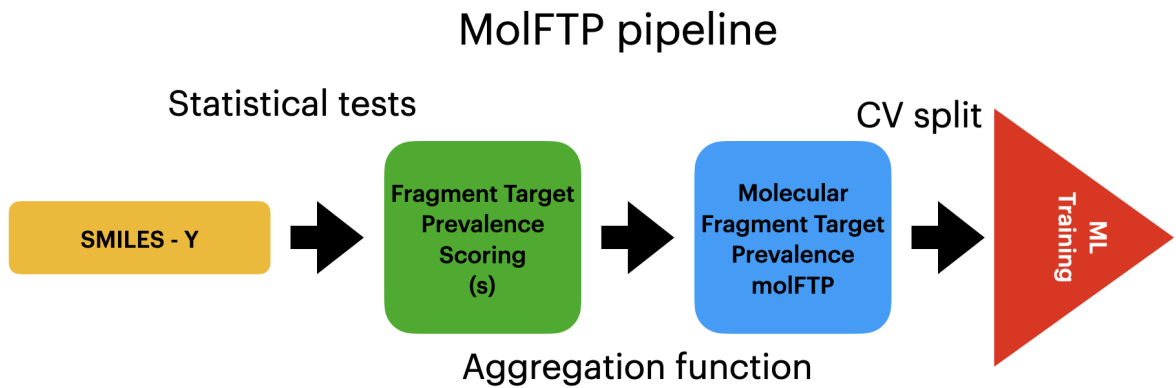

Figure F.3 : molFTP pipeline.

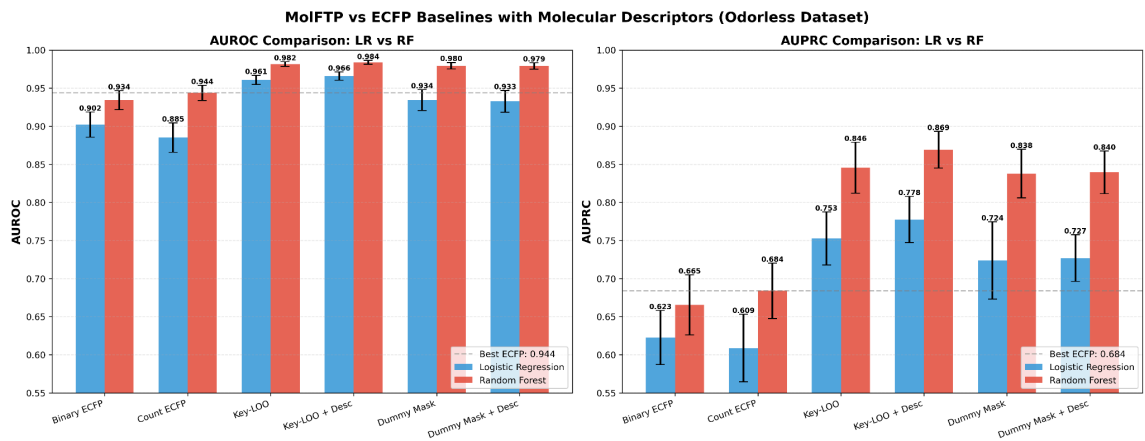

Figure F.4 comparison of baseline ECFP6 dim 2048, radius 3 with key-loo and dummy masking (with or without 3 extra molecular features) for logistic regression and random forest CV 10 AUPRC is clearly the best metric to compare in detailed model variations.

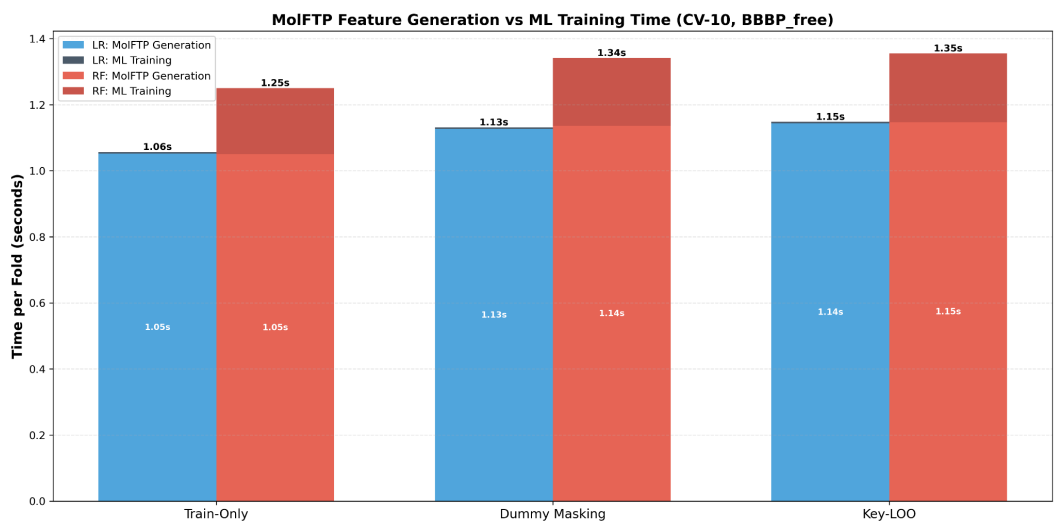

Figure F.5 Measure of runtime speed for generating molFTP and training a model (for aka 2000 molecules or 1800 for Train-only case).

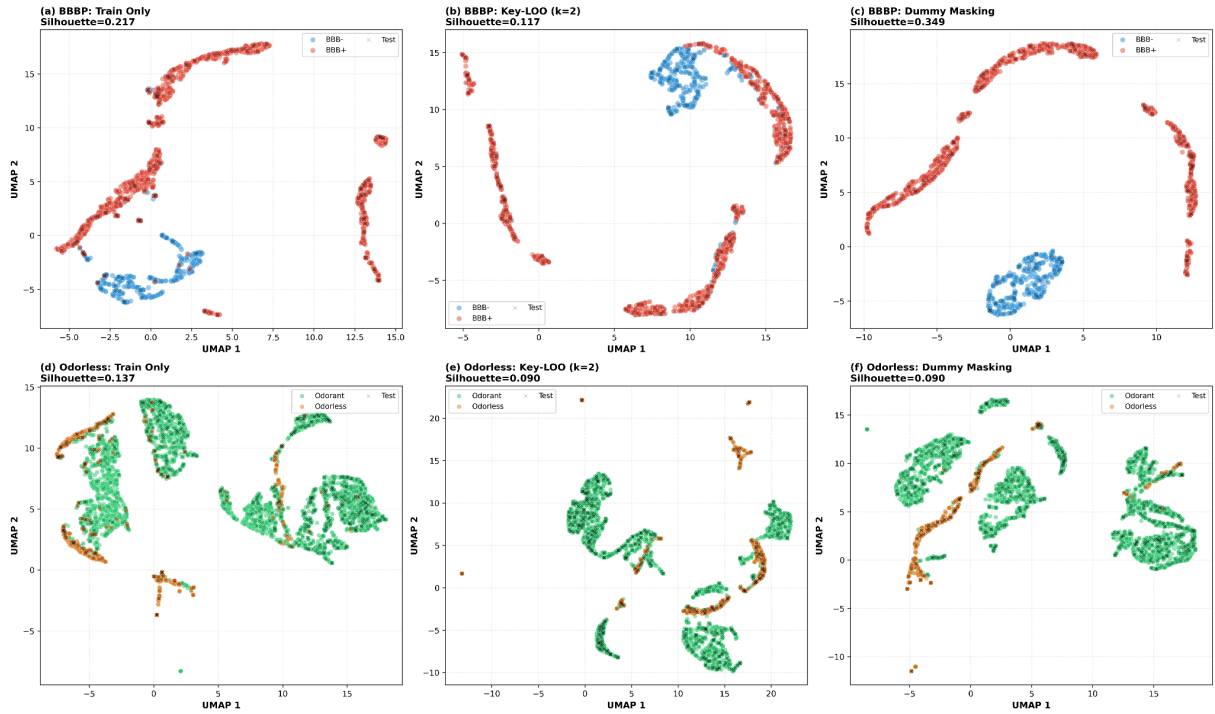

Figure F.6: UMAP BBBP_free and Odorless projections of the key-loo and masking compare to a train/test true split